\documentclass[letterpaper, 10 pt, conference]{ieeeconf}  

\IEEEoverridecommandlockouts                              
                                                          
\usepackage{color}
\usepackage{multirow}
\usepackage{booktabs} 
\usepackage{url}
\usepackage{subfigure}
\usepackage{cite}

\usepackage{float}
\usepackage{subfigure}
\usepackage[pdftex]{graphicx}

\usepackage{enumerate}
\usepackage{amssymb}
\usepackage{leftidx}
\usepackage{amsfonts}
\usepackage{amsmath}
\usepackage{bm}
\usepackage{booktabs}
\usepackage{setspace}
\usepackage{makecell}
\usepackage[colorlinks,linkcolor=blue]{hyperref}
\usepackage[ruled,linesnumbered]{algorithm2e}
\usepackage{mathrsfs}

\begin{document}
\title{Pole-like Objects Mapping and Long-Term Robot Localization in Dynamic Urban Scenarios}

\author{Zhihao Wang, Silin Li, Ming Cao, Haoyao Chen* and Yunhui Liu
	\thanks{This work was supported in part by the National Natural Science Foundation of China under Grant U1713206 and Grant 61673131. (Corresponding author: Haoyao Chen.)}
	\thanks{Z.H. Wang, S.L. Li, and H.Y. Chen* are with the School of Mechanical Engineering and Automation, Harbin Institute of Technology Shenzhen, P.R. China, e-mail: hychen5@hit.edu.cn.}
	\thanks{Y.H. Liu is with the Department of Mechanical and Automation Engineering, Chinese University of Hong Kong, P.R. China, e-mail: yhliu@mae.cuhk.edu.hk.}
}
\maketitle
\begin{abstract}
	Localization on 3D data is a challenging task for unmanned vehicles, especially in long-term dynamic urban scenarios.
	Due to the generality and long-term stability, the pole-like objects are very suitable as landmarks for unmanned vehicle localization in time-varing scenarios.
	In this paper, a long-term LiDAR-only localization algorithm based on semantic cluster map is proposed.
	At first, the Convolutional Neural Network(CNN) is used to infer the semantics of LiDAR point clouds. Combined with the point cloud segmentation, the long-term static objects pole/trunk in the scene are extracted and registered into a semantic cluster map. When the unmanned vehicle re-enters the environment again, the relocalization is completed by matching the clusters of the local map with the clusters of the global map. Furthermore,
	the continuous matching between the local and global maps stably outputs the global pose at 2Hz to correct the drift of the 3D LiDAR odometry. The proposed approach realizes localization in the long-term scenarios without maintaining the high-precision point cloud map.
  The experimental results on our campus dataset demonstrate that the proposed approach performs better in localization accuracy compared with the current state-of-the-art methods. The source of this paper is available at:
	\href{http://www.github.com/HITSZ-NRSL/long-term-localization}{http://www.github.com/HITSZ-NRSL/long-term-localization}.
\end{abstract} 
\IEEEpeerreviewmaketitle

\section{Introduction}
Simultaneous Localization and Mapping (SLAM) can provide location and map information for unmanned vehicles in unknown environments. Many classical SLAM methods \cite{murORB2, zhang2014loam, legoloam2018} have been proposed and applied in static environments. Because traditional point feature descriptors, such as SIFT, SURF and BRIEF, are robust to uniform changes in light intensity and slight changes in viewpoint.
However, the real urban environments are time-varying, such as, lighting changing, weathers and seasons varing, and the movement of objects (e.g., cars parked on both sides of the road may be driven away after a while), etc. And most descriptors do not take the large environmental changes into account. Thus, the dynamic environments introduce difficulties for a unmanned vehicle to localize itself or place recognition when it re-enters a known and mapped environment. Some literatures \cite{pomerleau2014long} rely on regularly updating high-definition point cloud maps to deal with the problem of dynamic environments, which is time and computation consuming.



To accomplish the long-term place recognition in dynamic environments, many approaches \cite{Churchill2013ExperiencebasedNF} tried to take multi-experiences into the localization framework. 
These approaches revealed inherent drawbacks because they need to capture various conditions for the same place as a prior to increase the size of the database of experiences. 
Contrast to visual appearance, the physical structure of a place rarely changes over time. Hence, utilizing structural information is beneficial for long-term localization \cite{Ye2017PlaceRI}. 
In that sense, a single observation using LiDAR could represent a canonical characteristic of a place, eliminating the need of multiple experiences for robust localization. 
In this line of research, handcrafted descriptor-based \cite{Kim2018ScanCE} and learning-based \cite{segmatch} methods for place recognition over point clouds have been widely proposed. 
However, these studies hardly captured the long-term localization requirements, including a slow but massive structural variance (e.g., construction and demolition) and unexpected viewpoint from the road topology change.

Inspired by the work of \cite{pole_localization}, pole-like objects, such as street lamps, poles of building and tree trunks, etc., 
are ubiquitous in urban areas; they are long-term stable and invariant under seasonal and weather changes, 
and their geometric shapes are well-defined. These advantages make pole-like objects suitable as landmarks for accurate and reliable relocalization. 
Thus, an approach based on semantic cluster is presented for long-term relocalization in urban dynamic environments, relying on pole/trunk landmarks extracted from mobile LiDAR data. 
We are not the first one to propose this kind of technique in long-term problem. However, to the best of our knowledge, this work is the first one that only uses 3D LiDAR to extract pole-like objects for place recognition and localization in long-term scenarios.
To summarize, the main contributions of this paper are three-fold.
\begin{itemize}
	\item A method to extract semantic cluster of pole-like objects from raw 3D LiDAR points and create a robust semantic cluster map is proposed to resolve the long-term challenge.
	\item A semantic cluster association algorithm based on geometric consistency is proposed to relocalize unmanned vehicles in long-term scenarios.
	\item A long-term and real-time localization system is developed based on the robust semantic cluster relocalization module. 

\end{itemize}

\section{Related Works}
Thanks to the rapid development of deep learning, some methods \cite{larsson2019cross} based on image semantic segmentation have been proposed to overcome environmental changes. 
Compared with the low-level point features, it is a more robust way to describe the environmental changes by using the high-level semantic features. 
Since geometric properties of the environments are typically more stable than its photometric characteristics, a geometry matching method \cite{caselitz2016matching} is proposed to achieve long-term camera localization by extracting object-level features in the environment.
Compared with the change of visual characteristics, the structure of the environment is stable, so many solutions based on laser sensors have been researched. Scan Context Image(SCI) \cite{Kim2019-SCI} is proposed to encode the top view of LiDAR point clouds into a three-channel color image, convert the place recognition problem to a classification problem.

Although many methods have been proposed for localization, few empirical studies showed the effectiveness of LiDAR descriptors on long-term localization capability in urban areas. As the high-level semantic information is robust to describe the environmental changes, 
a novel semantic graph \cite{kong2020semantic} is proposed for point clouds by reserving the semantic and topological information of the raw point clouds. Thus, place recognition is modeled as graph matching problem. However, this work has not been verified on the long-term place recognition problem. 
A global localization algorithm \cite{Ratz2020OneShotGL} is proposed by using only a single 3D LiDAR scan at a time, which relies on learning-based descriptors of point cloud segments. But its performance on long-term dataset is underperformance and not stable.

To address above problems, a long-term localization algorithm is proposed on the semantic cluster map of pole/trunks objects. 
The semantic cluster map is built by extracting the pole-like objects from raw 3D LiDAR points by CNN.
Then, the semantic cluster association algorithm based on geometric consistency vertification is proposed to obtain the pose transformation between the local semantic cluster map and the global semantic cluster map.
Finally, the long-term localization system utilized the pose transformation to correct the drift of LiDAR odometry in real time for high accuracy localization. 

\section{System Overview}
\begin{figure}[htbp]
  \centering
  \subfigure[Research platform]{\includegraphics[height=2.0in]{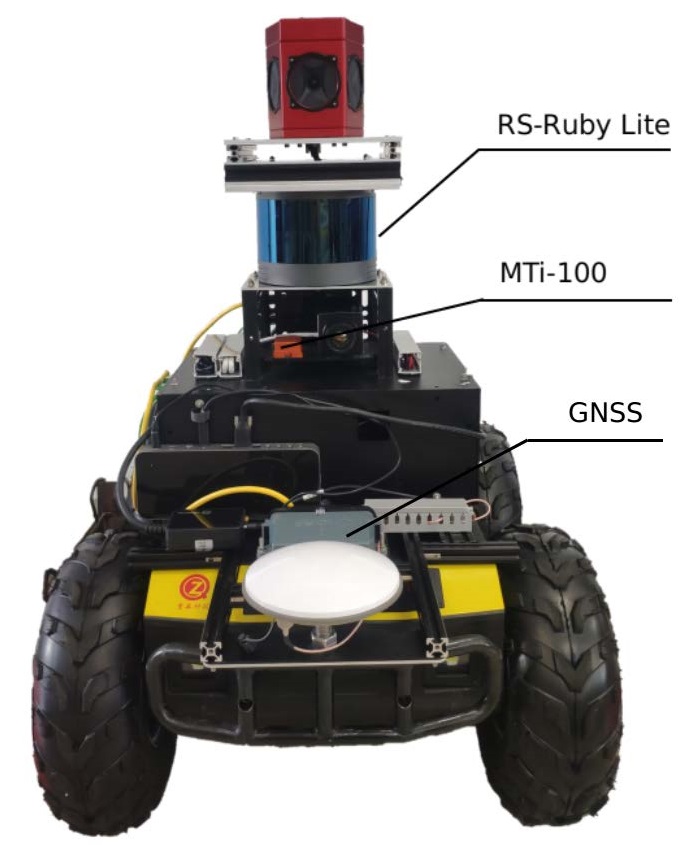}}
  \subfigure[System overview]{\includegraphics[height=2.0in]{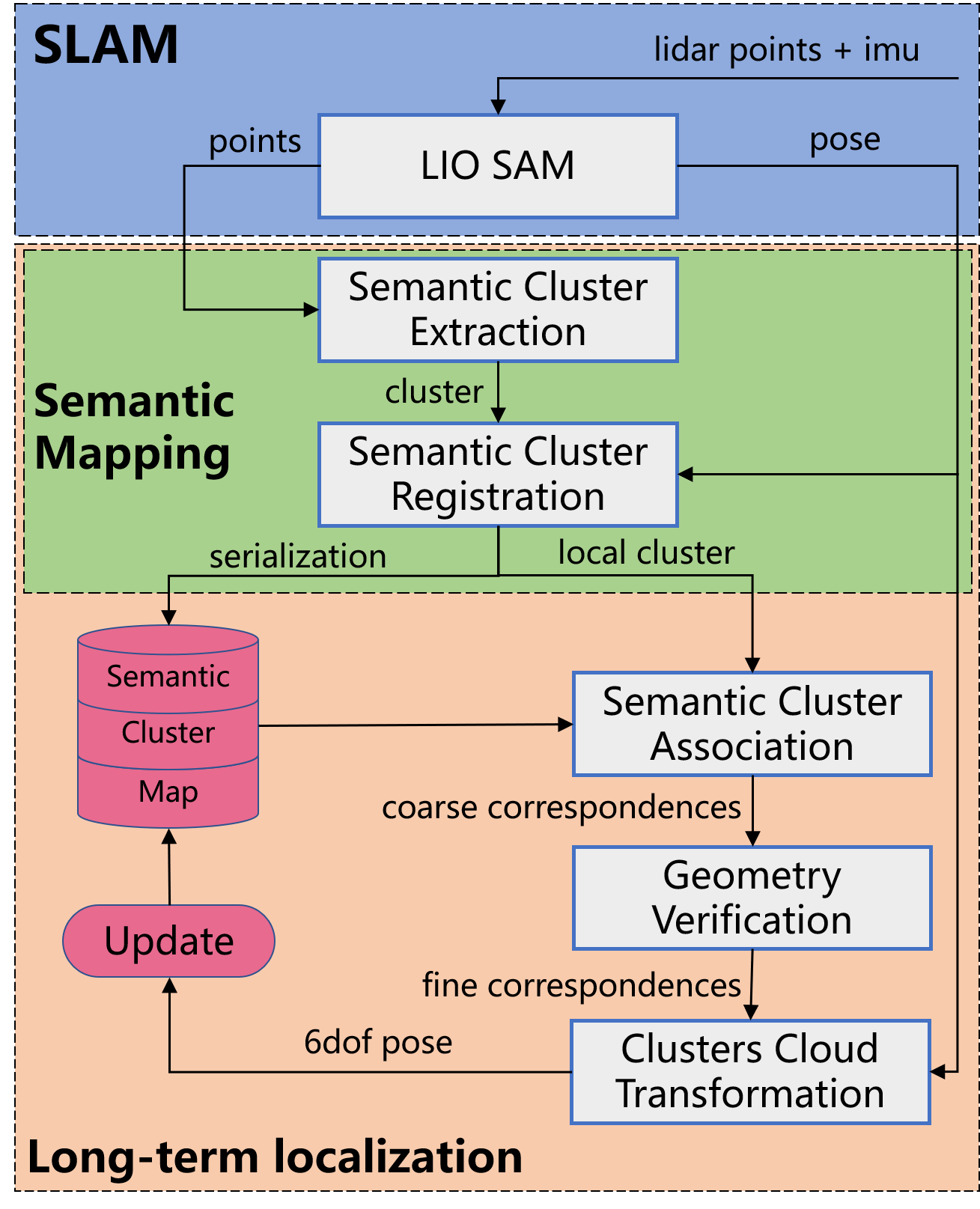}}
  \caption{Platform and system overview of long-term localization algorithm}
  \label{fig:System Overview}
\end{figure}

The experimental unmanned vehicle system is shown in Fig.\ref{fig:System Overview}(a). 
The RS-Ruby-Lite 3D LiDAR with 80 laser-beam has a vertical field of view (FOV) of $40^\circ$ and a horizontal FOV of $360^\circ$.
And the inertial sensor MTi-100 is also equipped with the unmanned vehicle MR1000.
Global Navigation Satellite System (GNSS)/Inertial Navigation System(INS) are used as the ground truth for comparison. The proposed system is validated on the laptop with Intel Core i7-7820HK CPU@2.90GHz on Robot Operating System (ROS).


The framework of the long-term localization system is shown as Fig.\ref{fig:System Overview}(b).
The semantic clusters of pole-like objects are extracted by the semantic lables of the point clouds, which are obtained directly through CNN.
Then, the semantic clusters of pole-like objects are registered in the global semantic cluster map using the six-degree-of-freedom(6-DOF) pose obtained by SLAM module.
When the unmanned vehicle re-enters the environment where the global semantic cluster map has been established, the local semantic cluster map is built by using the pose from SLAM module and the semantic clusters of the current frame. 
Then, a semantic cluster association algorithm based on the semantic labels and the geometric structure of semantic clusters is proposed to calculate the pose transformation between the local and the global semantic cluster map.
The relocalization module includes the semantic clusters registration and association. 
And the relocalization module continuously runs and stably outputs the global pose at 2Hz to correct the drift of the LiDAR odometry. The fine 6-DOF pose is published in real time after a drift correction. 
The high accuracy localization of the unmanned vehicle is realized in the long-term scenarios without maintaining a high precision point cloud map.

\section{Method}
\subsection{Semantic Cluster Extraction}
The convolutional neural network RangNet++ \cite{milioto2019iros} is used for semantic segmentation of point clouds. The semantic labels of LiDAR point clouds are annotated on self-made campus dataset and the network of RangNet++ is retrained for infering the semantic labels of point clouds.
The range-image based method\cite{bogoslavskyi16iros} is used to segment the point clouds into clusters based on the semantic labels. 
For each cluster, the label of cluster is voted by the statistical number of the point labels in the cluster, by considering the imperfection of point cloud semantic segmentation. 
In the end, two kinds of semantic clusters, i.e., trunk and pole, are obtained.

\begin{figure}[htpb]
	\centering
	\includegraphics[width=0.8\linewidth]{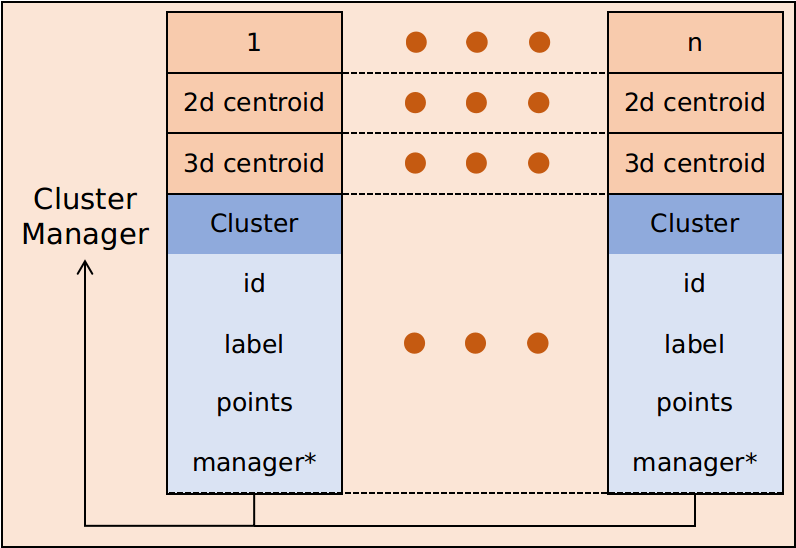}
	\caption{Memory structure of semantic cluster manager. The three dots in each row indicate continuous memory blocks. The 3d centroid is the average position of all points of a cluster in 3D, and the 2D centroid is obtained by projected the 3d centroid into XY plane. } 
	\label{fig:cluster_data_structure}
\end{figure}

Since the semantic cluster map and each semantic cluster need to be visited frequently, designing a semantic cluster manager is necessary to efficiently manage various attributes and operations of clusters. The manager includes 3D centroid points, 2D centroid points and several attributes, such as label, ID, point clouds and memory address. 
The memory structure is shown in the Fig.\ref{fig:cluster_data_structure}. 

\subsection{Semantic Cluster Registration}
The registration of semantic clusters will build a map, and the map includes two types: global semantic cluster map and local semantic cluster map. The global map represents the geometric information of pole-like objects in history environments, such as the environment of several months ago. The local map represents the geometric information of current environment.
As the global map need register the clusters of each frame from $t_0$ to $t_n$ into a global map, where $t_0$ and $t_n$ are the timestamps of the begin and end frame. The local semantic cluster map only uses the clusters of current frame at $t^{'}_k$ for association. This section only describes the registration of the global cluster map in details, while the registration of local map is similar.

Let $\mathcal{C}$ denote the semantic cluster, and
$\mathcal{M}^{cur}_{g}$ denote the global cluster map built on frames from $t_0$ to $t_{k-1}$. $\mathcal{M}^{cur}_{g}$ contains $j$ clusters, defined as
\begin{equation}
  \begin{aligned}
    \mathcal{M}^{cur}_g &= \{\mathcal{C}_{g1},\mathcal{C}_{g2},...,\mathcal{C}_{gj}\}
  \end{aligned}
  \label{equ:source_global_semantic_clusters_map}
\end{equation}
The global map is initialized by adding all the semantic clusters of the frame at $t_0$. 
After initialization, the clusters of current frame at $t_k$ need to be registered into the global map. 
Specifically, for each cluster $\mathcal{C}_{gk}$ in current frame at $t_k$, the position is transformed into the coordinate system of the global map by using LIO SAM \cite{liosam2020shan}.
The closest neighbor cluster of $\mathcal{C}_{gk}$ will be searched in the map $\mathcal{M}^{cur}_g$ using $k$ nearest neighbor(KNN) \cite{pcl}. And the label of $\mathcal{C}_{gk}$ is set as the same with the searched closest neighbor cluster. 
And if the closest neighbor cluster of $\mathcal{C}_{gk}$ is not found, $\mathcal{C}_{gk}$ is directly inserted into the global map as a new cluster $\mathcal{C}_{g \{j+1\}}$.

\begin{figure}[htpb]
	\centering
	\includegraphics[width=1.0\linewidth]{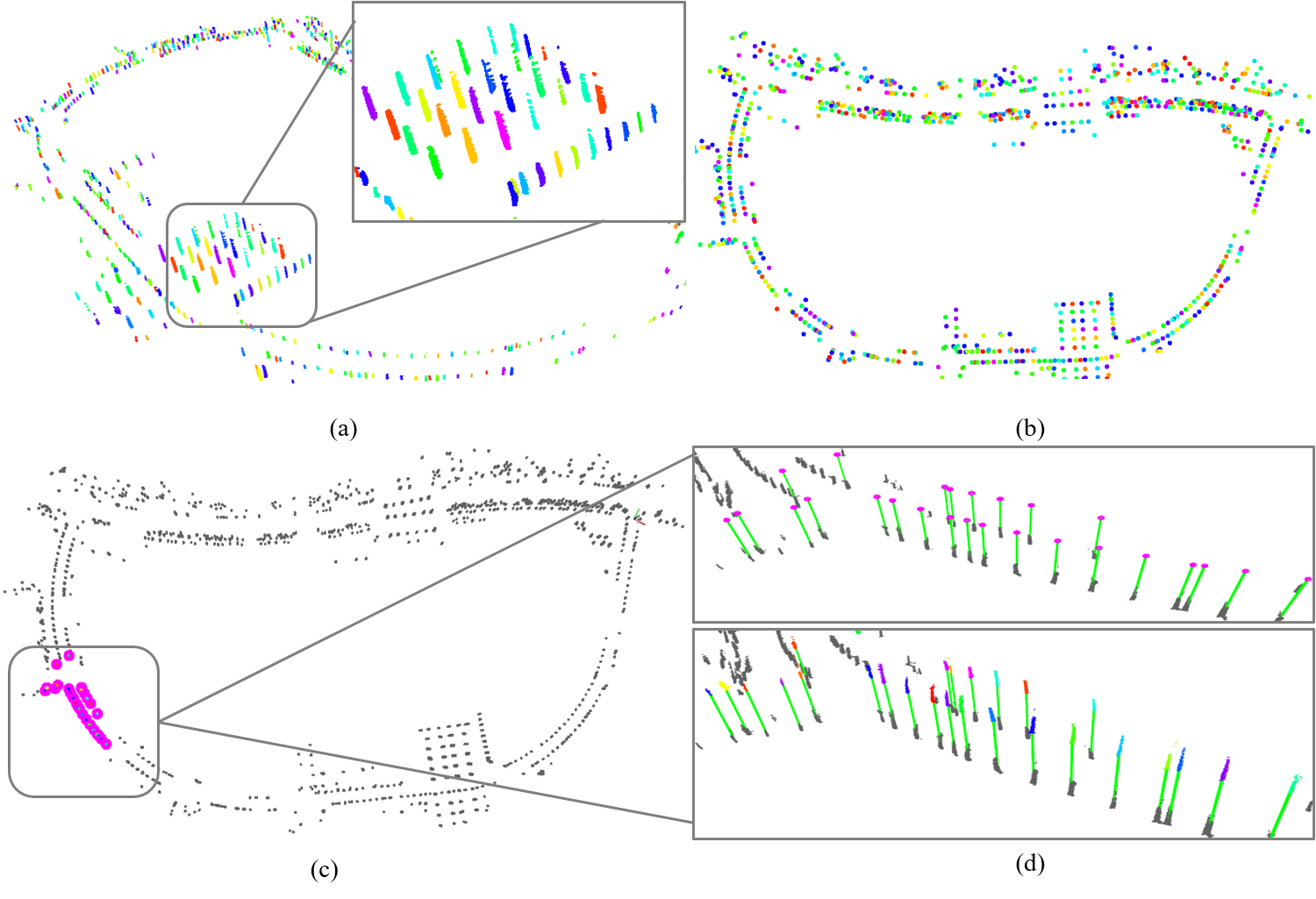}
	\caption{An illustration of the long-term relocalization approach. The 3D semantic cluster map is shown in (a), each color represents a semantic cluster. The 2D semantic cluster map is the top view of 3D cluster map in (b), and each color represents 3D centroid point of a semantic cluster. In (c), gray dots represent the clusters of the global semantic cluster map, pink dots represent the clusters of the local semantic cluster map. In (d), the green lines show the matched pairs between the local and global map. The upper figure shows the matched pairs of the 2D centroid points, while the lower figure shows the matched pairs of the 3D centroid points.}
	\label{fig:semantic_cluster_map}
\end{figure}

The registered 3D semantic cluster map is shown in Fig.\ref{fig:semantic_cluster_map}(a).
The clusters mainly include poles and trunks, and they are perpendicular to the horizontal plane in the environment. Therefore, the 2D centroid points are able to represent the geometric information of the clusters, and are registered into the cluster map. 
The registered 2D semantic cluster centroids are the XY-plane projection of 3D centroids, as shown in Fig.\ref{fig:semantic_cluster_map}(b). 

\subsection{Semantic Cluster Association}
The geometric distribution of neighbor clusters in a certain neighborhood around each cluster is utilized for clusters matching. 
The geometric information of the cluster to be matched is described as the relative positions and angles between the cluster and its neighbors. And all the geometric positions are simplified on the $XY$ plane for reducing computation cost.

Let $\mathcal{M}_{l}$ denote the local semantic cluster map of current frame with $P$ clusters.
The map registered by IV.B is loaded as the global semantic cluster map $\mathcal{M}_{g}$, which contains $Q$ clusters. 
And the definations are shown as follows:
\begin{equation}
  \begin{aligned}
    \mathcal{M}_l &= \{\mathcal{C}_{l1},\mathcal{C}_{l2},...,\mathcal{C}_{li},...,\mathcal{C}_{l{P}}\}\\
    \mathcal{M}_g &= \{\mathcal{C}_{g1},\mathcal{C}_{g2},...,\mathcal{C}_{gj},...,\mathcal{C}_{g{Q}}\}
  \end{aligned}
  \label{equ:global_semantic_clusters_map}
\end{equation}
In the local map $\mathcal{M}_{l}$, the neighbor clusters of $\mathcal{C}_{li}$ within the search radius $SR$ are denoted as $\mathcal{N}_{\mathcal{C}_{li}}$. 
Similarly, $\mathcal{N}_{\mathcal{C}_{gj}}$ denotes the neighbor clusters of $\mathcal{C}_{gj}$ within the search radius $SR$ in $\mathcal{M}_{g}$. And the number of neighbor clusters is $k_l$ and $k_g$, respectively. $\mathcal{N}_{\mathcal{C}_{li}}$ and $\mathcal{N}_{\mathcal{C}_{gj}}$ are denoted as follows:
\begin{equation}
  \begin{aligned}
    \mathcal{N}_{\mathcal{C}_{li}} &= RadiusSearch(\mathcal{M}_l, \mathcal{C}_{li}, SR)\\
																		&= \{\mathcal{C}_{li}^1,\mathcal{C}_{li}^2,...,\mathcal{C}_{li}^i,...,\mathcal{C}_{li}^{k_l}\} \\
    \mathcal{N}_{\mathcal{C}_{gj}} &= RadiusSearch(\mathcal{M}_g, \mathcal{C}_{gj}, SR)\\
                                &= \{\mathcal{C}_{gj}^1,\mathcal{C}_{gj}^2,...,\mathcal{C}_{gj}^j,...,\mathcal{C}_{gj}^{k_g}\} 
  \end{aligned}
  \label{equ:neighbor_clusters}
\end{equation}
The edges are the lines between $\mathcal{C}_{li}$ and its neighbor clusters in $\mathcal{N}_{\mathcal{C}_{li}}$, and these edges belong to the local cluster edge set ${E}_{\mathcal{C}_{li}}$.
Similarly, ${E}_{\mathcal{C}_{gj}}$ denotes the cluster edge set between $\mathcal{C}_{gj}$ and its neighbor clusters in $\mathcal{N}_{\mathcal{C}_{gj}}$. The expressions are shown as follows:
\begin{equation}
  \begin{aligned}
    {E}_{\mathcal{C}_{li}} &= \{e_l^1,e_l^2,...,e_l^i,...,e_l^{k_l}\}\\
    {E}_{\mathcal{C}_{gj}} &= \{e_g^1,e_g^2,...,e_g^j,...,e_g^{k_g}\}
  \end{aligned}
  \label{equ:neighbor_edges}
\end{equation}

The principles for matching two clusters are the same labels and enough matched edges. 
Then, the matching problem of semantic clusters is transformed into the matching of edges in the cluster edge sets.
$\mathcal{C}_{li}$ and $\mathcal{C}_{gj}$ are successfully matched if more than $N_{e}$ pairs of edges can be matched in ${E}_{\mathcal{C}_{li}}$ and ${E}_{\mathcal{C}_{gj}}$, $N_{e}$ denotes the minimum number of matched edges. Otherwise, they are not matched.

\subsubsection{Edge Association}
For the edge matching process of $e_l^i$ in ${E}_{\mathcal{C}_{li}}$, the length of each edge $e_g^j$ in ${E}_{\mathcal{C}_{gj}}$ are calculated, and the top $n$ edges whose length are nearest to the length of $e_l^i$ are selected as candidate edges of $e_l^i$, denoted as $\{{}^i\hat{e}_g^1,{}^i\hat{e}_g^2,...,{}^i\hat{e}_g^j,...,{}^i\hat{e}_g^n\}$, and $n$ is a constant(default: 5). 
$(e_l^i,{}^i\hat{e}_g^j)$ denotes the candidate edge pair, and the distance of the candidate edge pair is $\mathcal{D}_{li}^{gj}$, which will be defined later. Thus, the problem of edge matching is formulated as follows:


\begin{equation}
  \text{Min}\left(\mathcal{D}_{li}^{g1},\mathcal{D}_{li}^{g2},...,\mathcal{D}_{li}^{gj},...,\mathcal{D}_{li}^{gn} \right) < \delta_{e}
  \label{equ:edge_distance_condition}
\end{equation}
where $\delta_{e}$ is the maximum tolerance distance for successful matching of edge pairs. 

For matching $e_l^i$, not only the length and label of edge $e_l^i$ should be the same, but also the geometric relationships between the edge $e_l^i$ and the neighbors of $e_l^i$ should also be consistant. Therefore, sub-edge is introduced for verifying the neighbor geometric relationships.
The sub-edges of $e_l^i$ is the edges in edge set ${E}_{\mathcal{C}_{li}}$ except $e_l^i$, and composed the sub-edge set $SE_{li}$. 
Similarly, the edges in ${E}_{\mathcal{C}_{gj}}$ except ${}^i\hat{e}_g^j$ are called sub-edge set $SE_{gj}$ of ${}^i\hat{e}_g^j$:

\begin{equation}
  \begin{aligned}
    SE_{li} &= \{e_l^1,...,e_l^p,...,e_l^{k_l}\}, p \ne i \\
    SE_{gj} &= \{e_g^1,...,e_g^q,...,e_g^{k_g}\}, q \ne j
  \end{aligned}
  \label{equ:subedge_formular}
\end{equation}

The distance $\mathcal{D}_{li}^{gj}$ of the candidate edge pair $(e_l^i,{}^i\hat{e}_g^j)$ is defined as the average distances of successfully matched sub-edge pairs between the sub-edge sets $SE_{li}$ and $SE_{gj}$. 
And the matching of sub-edge pairs is in the following subsection. 

\subsubsection{Sub-Edge Association}

\begin{figure}[htpb]
	\centering
	\includegraphics[width=1.0\linewidth]{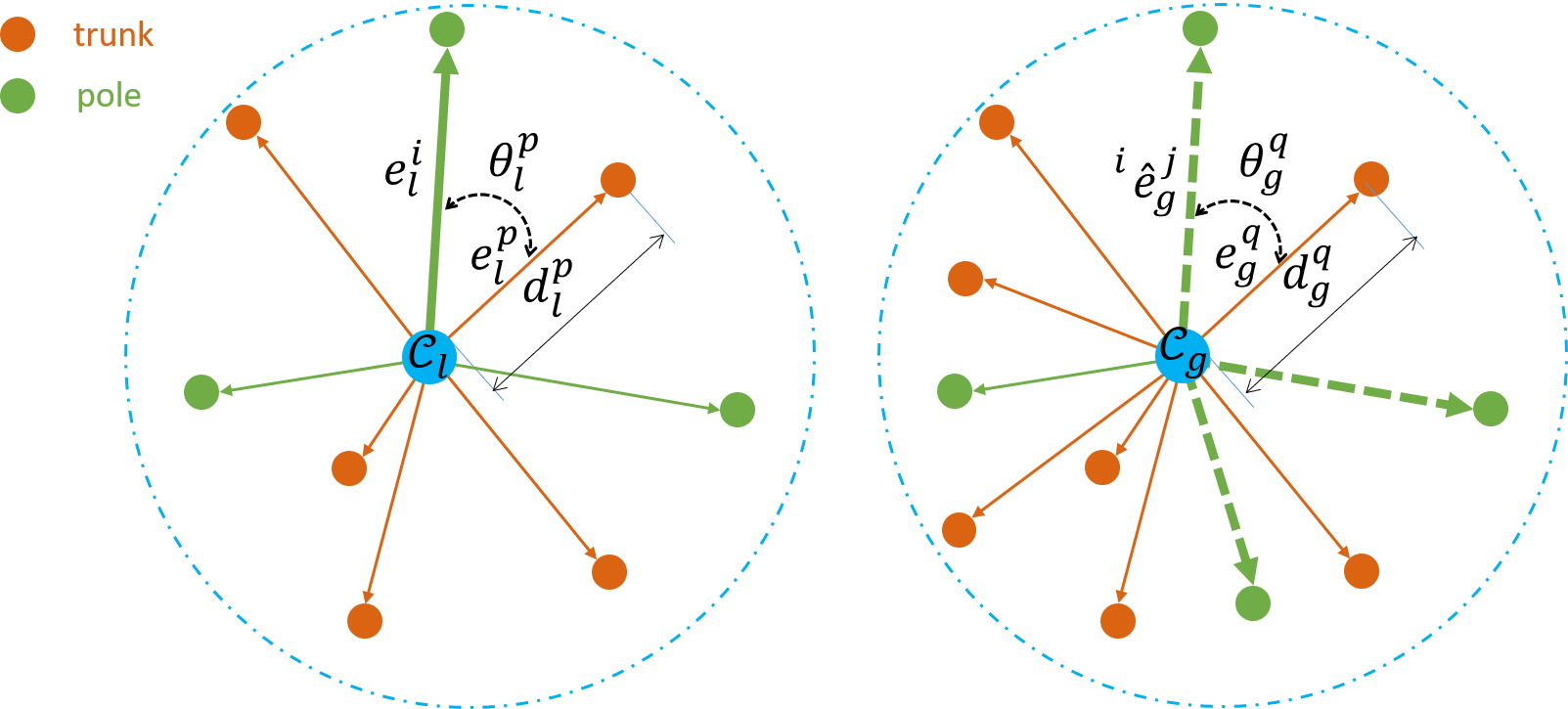}
	\caption{Cluster associaton diagram. (a) shows the neighbor semantic clusters of $\mathcal{C}_{li}$ in local semantic cluster map, and (b) shows the neighbors of $\mathcal{C}_{gj}$ in the global cluster map. Brown dots and green dots represent trunks and poles, respectively. Thick solid line with arrows in (a) represents the edge to be matched, and thin solid lines with arrows represent the sub-edges. $\theta, d$ represent the element of sub-edge. The thick dotted lines in (b) indicate the candidate edges of the edge to be matched in (a).
  } 
  \label{fig:cluster_association}
\end{figure}

The sub-edge $e_l^p$ of the matching edge $e_l^i$ in $SE_{li}$ includes elements $\left[d_l^p, \theta_l^p\right]$, where $d_l^p$ is the length of $e_l^p$, and $\theta_l^p$ is the clockwise angle between $e_l^p$ and $e_l^i$.
As the same, $\left[d_g^q, \theta_g^q\right]$ denotes the elements of sub-edge $e_g^q$ in $SE_{gj}$, where $d_g^q$ is the length of $e_g^q$, and $\theta_g^q$ is the clockwise angle between $e_g^q$ and $e_g^j$. 
A cluster association example is shown in Fig.\ref{fig:cluster_association}.

The matching success of two sub-edges $e_l^p$ and $e_g^q$ should satisfy four conditions: 
(a) the semantic cluster labels of two sub-edges are the same; 
(b) The length error between two sub-edges is under a certain threshold; 
(c) The angle error between two sub-edges is under a certain threshold; 
(d) The distance between two sub-edges, denoted as $d_{lp}^{gq}$, is also under a certain threshold. And the distance \cite{Pan2019} is defined as follows:

\begin{equation}
  d_{lp}^{gq} = \sqrt{d_l^p \cdot d_l^p+ d_g^q \cdot d_g^q - 2\cdot d_l^p\cdot d_g^q\cdot cos(|\theta_l^p-\theta_g^q|)}
  \label{equ:subedge_distance}
\end{equation}

The four conditions of sub-edges matching are mathematically formulated as follows: 

\begin{equation}
  \left\{
  \begin{aligned}
    Label(e_l^p) &== Label(e_g^q) \\
    |d_l^p - d_g^q| &< \delta_{d} \\
    |\theta_l^p-\theta_g^q| &< \delta_{\theta} \\
    d_{lp}^{gq} &< \delta_{se}
  \end{aligned}
  \right.
  \label{equ:subedge_match_conditions}
\end{equation}
where $Label()$ denotes the neighbor cluster label; 
$\delta_{d}$ indicates the threshold of distance error between sub-edges; 
$\delta_{\theta}$ indicates the threshold of the angle error of sub-edges; 
$\delta_{se}$ indicates the threshold of the distance of sub-edges.

\subsubsection{Semantic Cluster Association}
If more than $N_{se}$ sub-edges of the candidate edge pair $(e_l^i,{}^i\hat{e}_g^j)$ are satisfied with \eqref{equ:subedge_match_conditions}, the candidate edge pair is matched successfully. Then, the distance $\mathcal{D}_{li}^{gj}$ of candidate edge pair can be computed as the first line of \eqref{equ:edge_candidate_distance}. If the candidate edge pair is not matched, the distance $\mathcal{D}_{li}^{gj}$ is given infinite $DOUBLE\_MAX$.

\begin{equation}
  \mathcal{D}_{li}^{gj} = 
  \begin{cases}
    log\frac{k_l-1}{k_{se}^{ij}} \cdot \frac{1}{k_{se}^{ij}}\sum_{pq}^{k_{se}^{ij}} d_{lp}^{gq}, & k_{se}^{ij} >= N_{se} \\
    DOUBLE\_MAX, &\text{otherwise}
  \end{cases}
  \label{equ:edge_candidate_distance}
\end{equation}
where $k_{se}^{ij}$ indicates the number of sub-edge pairs successfully matched,
$k_l-1$ indicates the number of all sub-edges in $SE_{li}$, and $N_{se}$ is a threshold. 
As shown in \eqref{equ:edge_candidate_distance}, the distance $\mathcal{D}_{li}^{gj}$ considered the distances of all the matched sub-edges and the proportion of the matched sub-edges ${k_{se}^{ij}}$ to all sub-edges in $SE_{li}$. Considering these two factors is to avoid the false matching caused by small average distance of sub-edges and the low proportion of successful matched sub-edges pairs.

From \eqref{equ:edge_distance_condition} and \eqref{equ:edge_candidate_distance}, the best candidate edge pair is formulated as follows:

\begin{equation}
  \hat{\mathcal{D}_{li}} = \text{Min}\left(\mathcal{D}_{li}^{g1},\mathcal{D}_{li}^{g2},...,\mathcal{D}_{li}^{gj},...,\mathcal{D}_{li}^{gn} \right) < \delta_{e}
  \label{equ:edge_distance_condition_total}
\end{equation}
where $\hat{\mathcal{D}_{li}}$ is the distance of the best candidate edge pair, $\delta_{e}$ denotes the threshold of the distance.

After matched the candidate edge pairs, the labels and the matched edge pairs numbers of the two semantic clusters to be matched should also be considered. Thus, the matching of clusters are denoted as follows:

\begin{equation}
  \begin{aligned}
    Label(\mathcal{C}_{li}) &== Label(\mathcal{C}_{gj}) \\
    k_e &>= N_{e}    
  \end{aligned}
  \label{equ:cluster_match_condition}
\end{equation}
where $k_e$ is the matched edges pairs numbers, and $N_{e}$ represents the minimum number of edge pairs successfully matched. 
Algorithm \ref{algo:semantic_cluster_association} shows the algorithm of semantic cluster association. 
An matching result example of semantic cluster association is shown in Fig.\ref{fig:semantic_cluster_map}(c) and (d).

\begin{algorithm}
	\caption{Semantic Cluster Association}
	\label{algo:semantic_cluster_association}
	\KwIn{Two clusters that need to be matched $\mathcal{C}_{li}$, $\mathcal{C}_{gj}$ and their respective cluster map $\mathcal{M}_l$, $\mathcal{M}_g$}
	\KwOut{Result of the two clusters associated \text{True/False}}
  \If{$Label(\mathcal{C}_{li}) \ne Label(\mathcal{C}_{gj})$}{
    return False;
  }
  Search nearest neighbor clusters and get neighbor edges ${E}_l, {E}_g$ of $\mathcal{C}_{li}$, $\mathcal{C}_{gj}$ by \eqref{equ:neighbor_clusters} and \eqref{equ:neighbor_edges}.

	$k_{e} \leftarrow 0$
																																																							
	\For {$e_l^i : {E}_l$} {
		$candidates = $ FindNEdgePairCandidates$({E}_g, e_l^i)$ \\
		\For {$ \left(e_l^i, {}^i\hat{e}_g^j\right) : candidates$}{

			$k_{se}^{ij} \leftarrow 0$ \\

			\For{$e_l^p : SE_{li}$}{
				\For{$e_g^q: SE_{gj}$}{
          Compute sub-edge matching result $IsSubEdgeMatch$ by \eqref{equ:subedge_match_conditions} and \eqref{equ:subedge_distance}\\
          \If{$IsSubEdgeMatch$ is True}{
            $k_{se}^{ij} \leftarrow k_{se}^{ij}+1$ 
          }
				}
			}				
      Compute edge pair distance $\mathcal{D}_{li}^{gj}$ by \eqref{equ:edge_candidate_distance}\\
    }

    Compute matching result $IsEdgeMatch$ by \eqref{equ:edge_distance_condition_total} \\

		\If{$IsEdgeMatch$ is True}{
			$k_{e} \leftarrow k_{e} + 1$
		}
  }
	return $k_{e} >= N_e$ (\eqref{equ:cluster_match_condition})
\end{algorithm}

\subsection{Long-term Relocalization}
Algorithm \ref{algo:semantic_cluster_association} is utilized to match the semantic clusters in the local map with the global map, and provides $n$ pairs of semantic cluster matched pairs. Let $c_i = (\mathcal{C}_{li}, \mathcal{C}_{gi})$ denote a matched pair of semantic clusters, where $\mathcal{C}_{li} \in \mathcal{M}_l$ and $\mathcal{C}_{gi} \in \mathcal{M}_g$. 
Semantic cluster matching pairs obtained from semantic cluster association algorithm are coarse correspondences. 
Thus, a geometric consistency method \cite{gollub2017partitioned} will be used to eliminate the false positive matched pairs, and finally fine correspondences are remained.

For two matched pairs $(c_i,c_j)$, if their euclidean distance error in the local and global map is under a certain threshold, $(c_i,c_j)$ is geometric consistent, formulated as follows:
\begin{equation}
  \left|d_{l}\left(c_{i}, c_{j}\right)-d_{g}\left(c_{i}, c_{j}\right)\right| \leq \epsilon
  \label{equ:geometric_consistant}
\end{equation}
where $\epsilon$ indicates the maximum tolerance euclidean distance error; 
$d_{l}$ represents the euclidean distances of $\mathcal{C}_{li}$ and $\mathcal{C}_{s_j}$ in $\mathcal{M}_l$,
and $d_{g}$ represents the euclidean distances of $\mathcal{C}_{gi}$ and $\mathcal{C}_{gj}$ in $\mathcal{M}_g$.

Then, the 6-DOF pose transformation between the local map $\mathcal{M}_l$ and the global map $\mathcal{M}_g$ is computed by matching the corresponding semantic cluster centroid points and semantic cluster point clouds.
Firstly, the 3D centroid points pairs are extracted from semantic cluster matched pairs, and the RANSAC\cite{derpanis2010overview} algorithm is used to filter out wrong matched pairs.
Secondly, the coarse coordinate transformation matrix $\mathbf{{T}}_{gl}$ from the local map to the global map is calculated by using iterative closest point (ICP) algorithm with \eqref{equ:transformation}.
\begin{equation}
	\mathbf{{T}}_{gl}=\mathop{\arg\min}_{\mathbf{{T}}_{gl}} \frac{1}{2} \sum^{k}\left\|O_{li}-\mathbf{{T}}_{gl} O_{gj}\right\|^{2}  
	\label{equ:transformation}
\end{equation}
where $k$ is the number of the filtered semantic cluster matched pairs. $O_{li}$ and $O_{gj}$ are the 3D centroid points of the clusters $\mathcal{C}_{li}$ and $\mathcal{C}_{gj}$.

Thirdly, the local and global point clouds of the matched cluster pairs are respectively extracted from cluster manager. 
Using the coarse pose transformation $\mathbf{{T}}_{gl}$ as the initial value, the fine transformation matrix $\mathbf{\widehat{T}}_{gl}$ is obtained by the ICP algorithm again with the extracted point clouds instead of the 3D centroid points, shown as follows:
\begin{equation}
	\mathbf{\widehat{T}}_{gl}=\mathop{\arg\min}_{\mathbf{\widehat{T}}_{gl}} \frac{1}{2} \sum^{k}\left\|P_{li}-\mathbf{\widehat{T}}_{gl} P_{gj}\right\|^{2}  
	\label{equ:fine transformation}
\end{equation}
where $P_{li}$, $P_{gj}$ are the extracted point clouds of the clusters $\mathcal{C}_{li}$ and $\mathcal{C}_{gj}$; $\mathbf{\widehat{T}}_{gl}$ represents the global pose of the local map in the global map.

\subsection{Long-term Localization}

\begin{figure}[htpb]
  \centering
  \includegraphics[width=0.36\textwidth]{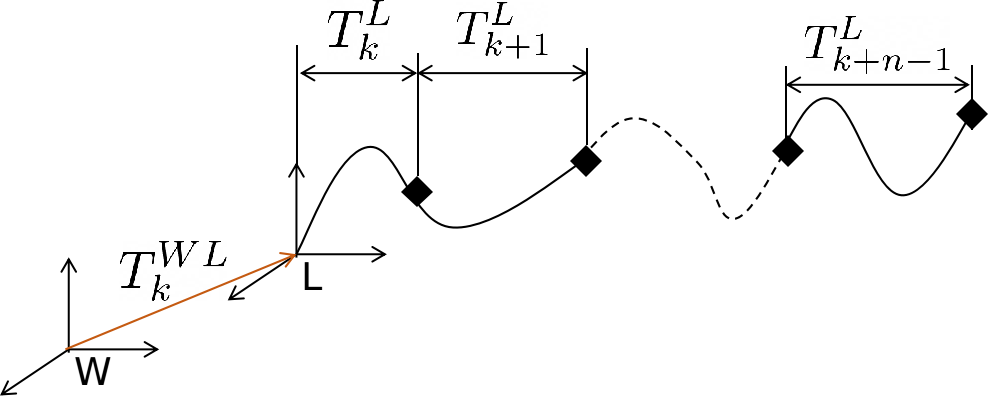}
  \caption{long-term localization process. $\{W\}$ denotes the world frame, $\{L\}$ denotes the LiDAR frame.}
  \label{fig:localization_formular}
\end{figure}

The long-term localization algorithm is shown in the Fig.\ref{fig:localization_formular}, the global pose $\mathbf{\widehat{T}}_{gl}$, stably output at 2Hz by the relocalization module, is used to correct the accumulated drift of the LiDAR odometry in real time.
The global pose $\mathbf{\widehat{T}}_{gl}$ at $t_k$ is re-expressed as $T_k^{WL}$. The LiDAR odometry outputs $n$ poses continuously from $t_k$ to $t_{k+n}$ before the next global pose is updated, denoted as $\{T_k^L, T_{k+1}^L,...,T_{k+i}^L,...,T_{k+n-1}^L\}$.
During $t_k$ to $t_{k+n}$, the global pose estimation of the unmanned vehicle is calculated as follows: 

\begin{equation}
  T_{k+i}^{WL} = T_k^{WL} \cdot {\prod}_{j=0}^{i} T_{k+j}^L
  \label{equ:lidar_global_pose}
\end{equation}
where $i \in [0,n-1]$.
So far, long-term localization has been finished by correcting the drift of LiDAR odometry with the real-time global pose of relocalization module.

\section{Experiements}
To evaluate the proposed relocalization and localization algorithms in long-term scenarios, several experiments were performed on the self-made campus dataset. 
It is worth noted that Carlevaris-Bianco\cite{carlevaris2016university} proposed NCLT dataset in long-term scene, but the Velodyne32 LiDAR in NCLT dataset is installed upside down, which makes the actual LiDAR point cloud sparsity equal to Velodyne16 LiDAR. While the proposed method is based on semantic segmentation of dense LiDAR point clouds, NCLT dataset cannot meet our requirements. 
Therefore, the proposed method is compared on the self-made dataset with the SCI algorithm\cite{Kim2019-SCI}, which is the state-of-the-art long-term LiDAR-based localization system.

\subsection{Benchmark Dataset}

\begin{figure}[htpb]
  \centering
  \includegraphics[width=0.4\textwidth]{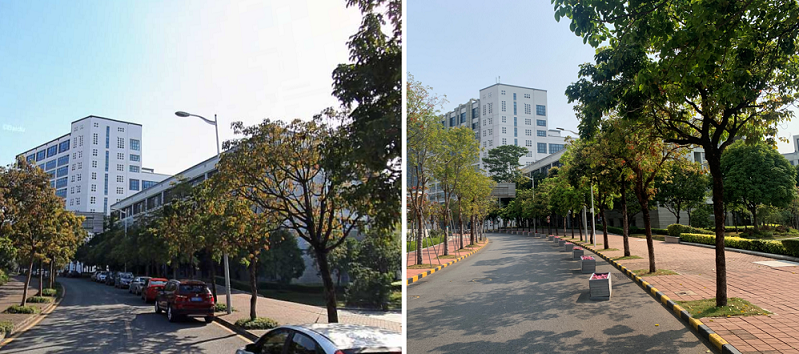}
  \caption{Dataset Environment. The left image is several months earlier than the right image, the differences are the removed cars, the trees with different shapes, and the lighting conditions.}
  \label{fig:university_town_environment}
\end{figure}
The challenge of long-term localization is the environmental changes. So the self-made dataset recorded the data in the university town environment with high traffic volume and dense vegetation.
The environment is shown in Fig.\ref{fig:university_town_environment}.

\begin{figure}[htpb]
  \centering
  \includegraphics[width=0.36\textwidth]{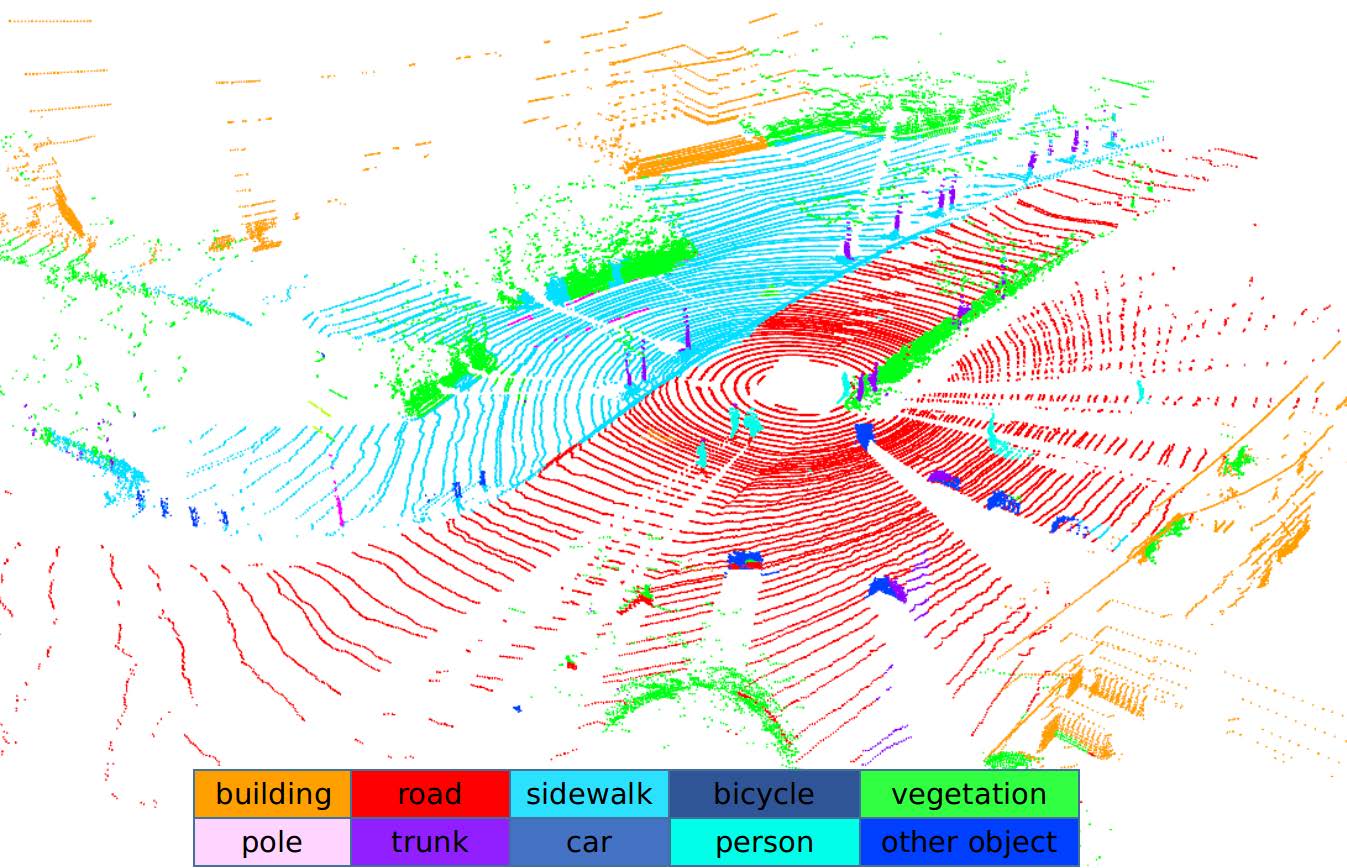}
  \caption{Example of annotated semantic points.}
  \label{fig:annotated_example}
\end{figure}

To use RangeNet++ to infer the semantic label, the self-made dataset is annotated with more than 1500 frames of 3D point clouds generated by LiDAR scanning, which covered a geographic area of 76,000 $m^2$ and included 120 million three-dimensional semantic points. 
And the marked labels with 10 categories included buildings, highways, sidewalks, bicycles, green plants, poles, tree trunks, vehicles, pedestrians, and other objects. 
An example of the annotated semantic point clouds by the labeling tool PointLabeler\cite{behley2019semantickitti} is shown in Fig.\ref{fig:annotated_example}. 
The organization of the semantic point clouds is labeled and organized according to the SemanticKitti\cite{behley2019semantickitti}. Four datasets were collected, and Table \ref{table:List of campus data sets} showed the details of datasets. Although the datasets were collected within a month, some places of the environment have changed significantly due to the movement of cars and growing of trees; furthermore, different numbers of clusters in the global map were randomly removed to simulate the significant changes of environment. We will also continue collect more datasets and update the results in the future.

\begin{table}[htpb]
  \caption{List of campus data sets.}
  \begin{center}
    \begin{tabular}{cccc}
      \hline
    Date       & Length(km) & Time      & Weather \\ \hline
    2020-10-12 & 1.31   & Afternoon & Sunny  \\ \hline
    2020-10-18 & 1.28   & Morning   & Cloudy  \\ \hline
    2020-10-25 & 1.53   & Evening   & Cloudy   \\ \hline
    2020-11-05 & 1.4    & Afternoon   & Sunny  \\ \hline
    \end{tabular}
  \end{center}  
  \label{table:List of campus data sets}
\end{table}


\subsection{Evaluation of Long-term Relocalization}
To evaluate the success rate of relocalization in long-term scenarios, the earliest dataset 2020-10-12 was used to build the global semantic cluster map, and the other three datasets were used to create the local semantic cluster maps for relocalization with the global map. 
The defination of successful relocalization is expressed as follows:
\begin{equation}
	\left\|t_{e s t}-t_{g t}\right\|<\delta_{\text {relocalization }} 
	\label{equ:success rate}
\end{equation}
where $t_{e s t}$ and $t_{g t}$ represent the estimated position of relocalization and the ground truth by GNSS; $\delta_{\text {relocalization }} $ denotes the maximum tolerance distance between the estimated position and the ground truth when the relocalization is successful, and set to 10 m. 

\begin{figure}[htpb]
  \centering
  \includegraphics[width=0.4\textwidth]{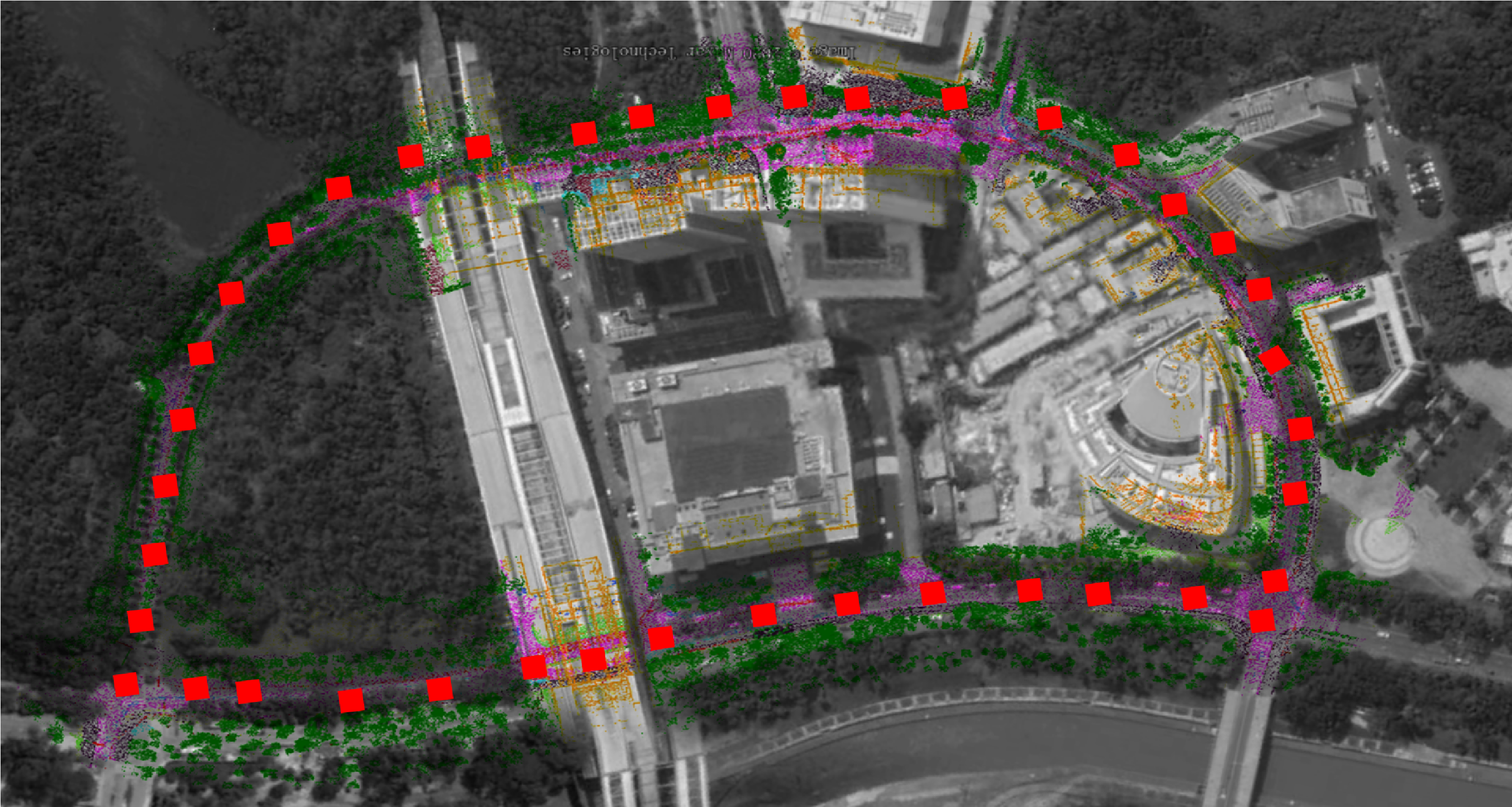}
  \caption{The positions distribution of the relocalization experiments. Red squares represent the selected positions.}
  \label{fig:approximate location distribution}
\end{figure}

The success rate of relocalization of the SCI algorithm are divided into two categories, namely the success rate of the highest(top 1) scoring frame and the top 5 scoring frames. 
120 positions were selected for the relocalization experiments on each dataset, and distributed as shown in Fig.\ref{fig:approximate location distribution}. And the parameters of clusters registration and association are shown in Table \ref{table:parameters analysis}.
And the success rate of relocalization was shown in Table \ref{table:Success rate Comparison}. 
The left number of each cell represents the number of successful relocalizations in 120 experiments, and the right number of each cell represents the success rate. 
The results showed the success rate of the proposed relocalization algorithm higher than the SCI algorithm on the self-made dataset.

\begin{table}[htpb]
  \caption{The parameters of clusters registration and association}
  \begin{center}
    \resizebox{\columnwidth}{!}{
      \begin{tabular}{cccc}
        \hline
      Parameter       & Description & Setting      \\ \hline
      SR & search radius    & 50 m   \\ \hline
      $\delta_{d}$& threshold of subedge distance error  & 0.3 m   \\ \hline
      $\delta_{\theta}$& threshold of subedge angle error  & $10^\circ$   \\ \hline
      $\delta_{se}$& threshold of subedge distance  & 0.2 m   \\ \hline
      $\delta_{e}$& threshold of subedge distance  & 0.25 m   \\ \hline
      $N_{se}$& minimum matched pairs of subedges  & 5   \\ \hline
      $N_{e}$& minimum matched pairs of edges  & 5   \\ \hline
      \end{tabular}
    }
  \end{center}  
  \label{table:parameters analysis}
\end{table}

\begin{table}[htpb]
  \caption{The Success rate Comparison of SCI and the proposed method.}
  \begin{center}
    \begin{tabular}{cccc}
      \hline
    Dataset       & SCI(TOP1) & SCI(TOP5)      & Proposed \\ \hline
    2020-10-18 & 99/82.5$\%$   & 114/95.0$\%$   & \bf{120/100.0$\%$}  \\ \hline
    2020-10-25 & 97/80.8$\%$   & 108/90.0$\%$   & \bf{116/96.6$\%$}   \\ \hline
    2020-11-05 & 103/85.8$\%$   & 112/93.3$\%$   & \bf{118/98.3$\%$}  \\ \hline
    \end{tabular}
  \end{center}  
  \label{table:Success rate Comparison}
\end{table}

\begin{figure}[htbp]
  \centering
  \subfigure[]{\includegraphics[height=1.2in]{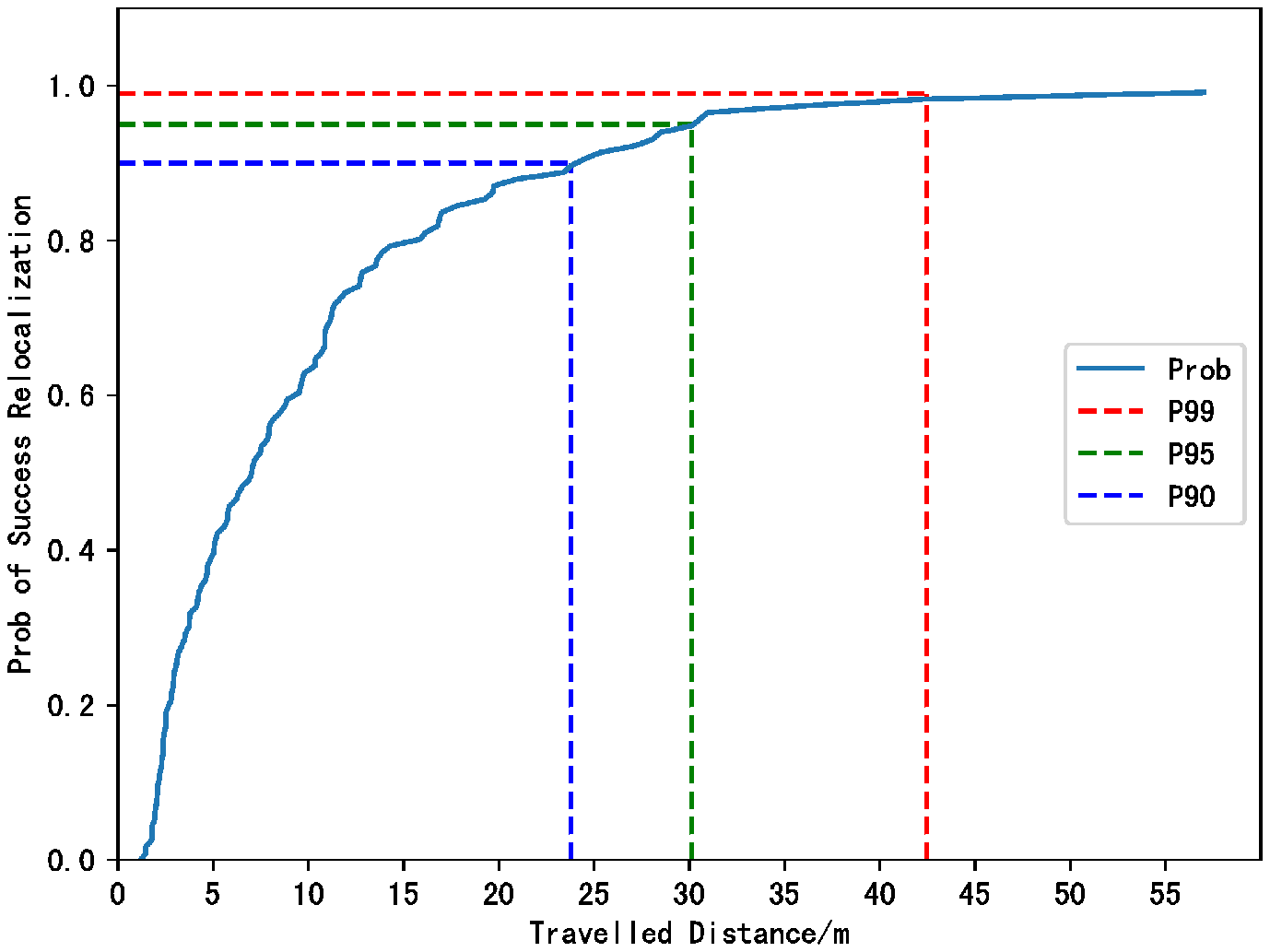}}
  \subfigure[]{\includegraphics[height=1.2in]{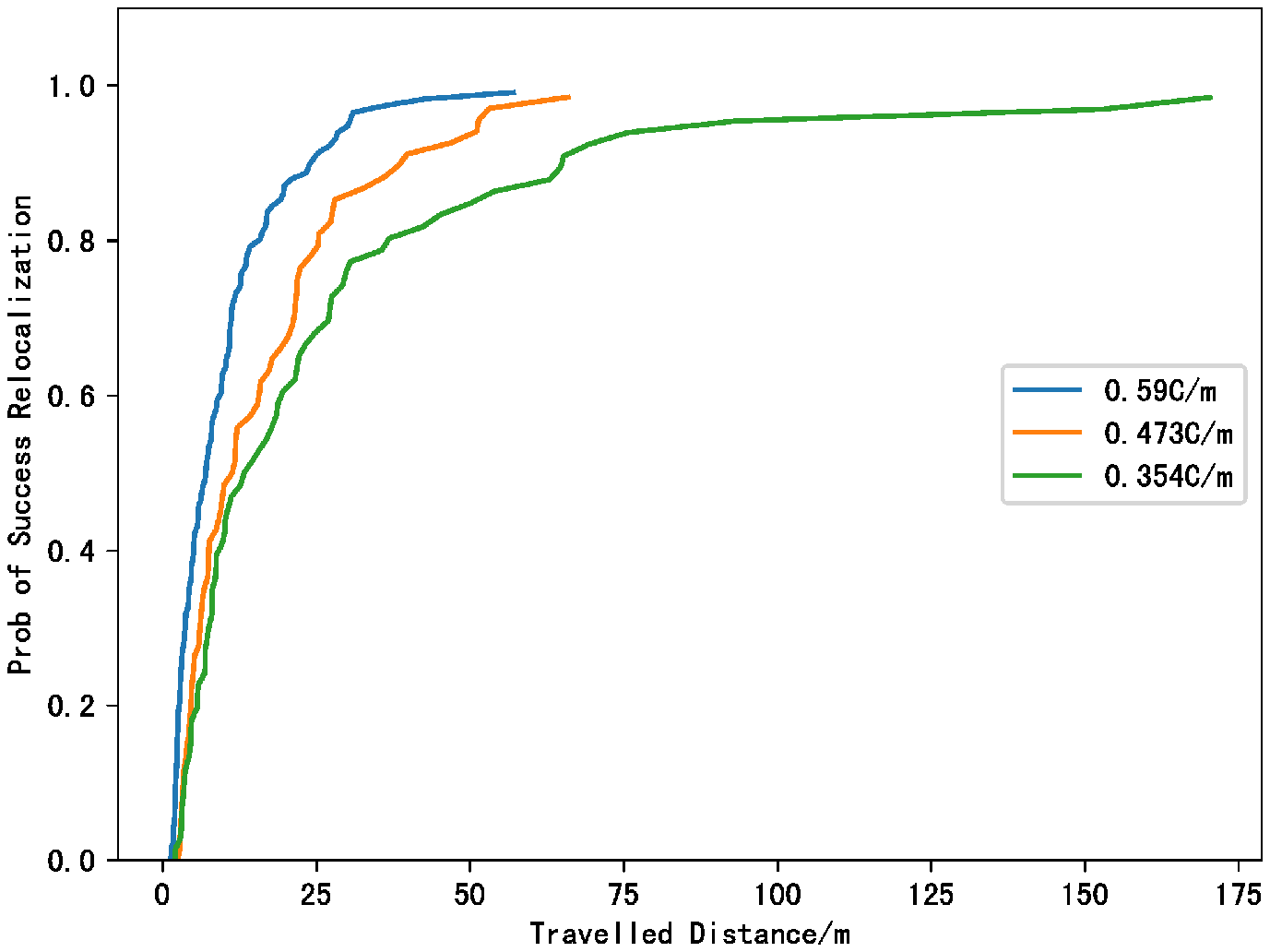}}
  \caption{(a) The result of relocalization success probability vs the travelled distance. The cyan solid curve denotes the relocalization success probability, and the dot lines denote different probability metrics. (b) The result of relocalization success probability vs the travelled distance under different cluster densities. C/m is the unit of density, denoting the number of semantic clusters of per meter.}
  \label{fig:success rate and distance}
\end{figure}

Moerover, to vertify how far unmanned vehicles have travelled when relocalization is successful, the travelled distance of successful relocalization were collected and sorted from 116 experiments, and the probability curve result is shown in Fig.\ref{fig:success rate and distance}(a).
When the unmanned vehicles travelled 24 m, the relocalization is done successfully with the probability of 90$\%$; 
And with the success probability of 95$\%$ and 99$\%$ when the unmanned vehicles travelled 30 m and 43 m.
The result illustrated that, by using the proposed relocalization algorithm in the long-term scenarios, the travelled distance of successful relocalization is relatively short compared with the scale of the whole map(almost 1.4 km).

To evaluate the performance of the proposed relocalization algorithm under large changes in the environment, different numbers clusters in the global semantic map were randomly selected to associate with local semantic cluster maps. 
Specifically, 827 semantic clusters were extracted from the built semantic map using the dataset 2020-10-12. When using the dataset 2020-11-05 for relocalization, a series of sub-global semantic cluster maps were randomly created by retaining 100$\%$, 80$\%$ and 60$\%$ of the total clusters of the global map by dataset 2020-10-12, i.e., 827, 661 and 496 clusters retained, respectively. 
And the relocalization experiments were compared between above three sub-global maps with different cluster numbers by the dataset 2020-10-12 and the local cluster maps by the dataset 2020-11-05.
The number of semantic clusters of per meter along the unmanned vehicle trajectory is defined as the semantic clustering density $\rho$, formulated as follows:

\begin{equation}
  \rho = {N}/{L}
  \label{equ:cluster density}
\end{equation}
where $N$ denotes the number of semantic clusters used in global map, and $L$ denotes the travelled distance.

For the three sub-global maps, the semantic cluster densities are calculated by \eqref{equ:cluster density} to be 0.59, 0.473 and 0.354 Clusters/m, respectively. 116, 68, and 62 times of the experiment were performed on the dataset 2020-11-05 to calculate the travelled distance when successfully relocated, the result was shown in Fig.\ref{fig:success rate and distance}(b). 
It can be seen that the larger density of semantic cluster map brings the shorter travelled distance. 
And when the semantic clustering density is not less than 0.473 Clusters/m, the unmanned vehicle only needs to travel no more than 50 m when the relocalization is successful with probability 90$\%$.

\subsection{Long-term Localization Analysis}
The evaluation of long-term localization experiment is mainly to compare the position estimation accuracy of the LiDAR odometry. 
The dataset 2020-10-05 was used to build the global semantic cluster map, and the other three datasets were used to create local semantic cluster maps for long-term localization.
The metrics of Root Mean Square Error(RMSE) is used for evaluation, and the results of the proposed long-term localization system and LOAM are compared with the ground truth.
The results were shown in Table \ref{table:RMSE of long-term localization}. The proposed approach does not need to utilize the IMU data for relocalization and localization, and therefore it was compared with the LOAM but not LIO-SAM. 

\begin{table}[htpb]
  \caption{RMSE of long-term localization.}
  \begin{center}
    \begin{tabular}{cccc}
      \hline
    Dataset       & LOAM(m)      & Proposed(m) \\ \hline
    2020-10-18 & 42.41      & \bf{0.29}  \\ \hline
    2020-10-25 & 44.32     & \bf{0.34}   \\ \hline
    2020-11-05 & 40.48     & \bf{0.24}  \\ \hline
    \end{tabular}
  \end{center}  
  \label{table:RMSE of long-term localization}
\end{table}

\begin{figure}[htbp]
  \centering
  \subfigure[]{\includegraphics[height=1.65in]{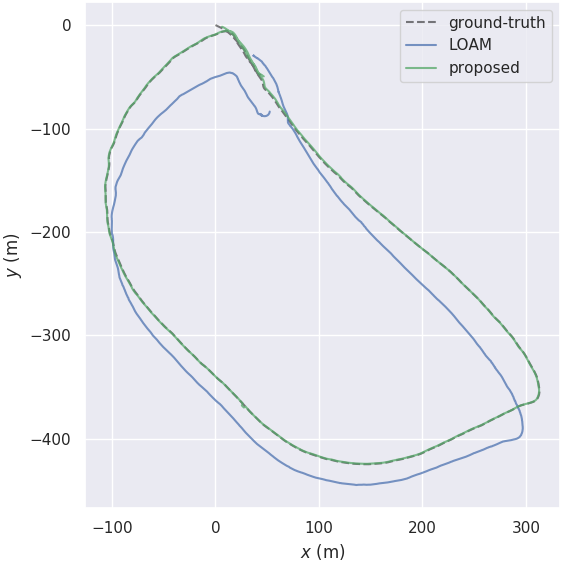}}
  \subfigure[]{\includegraphics[height=1.65in]{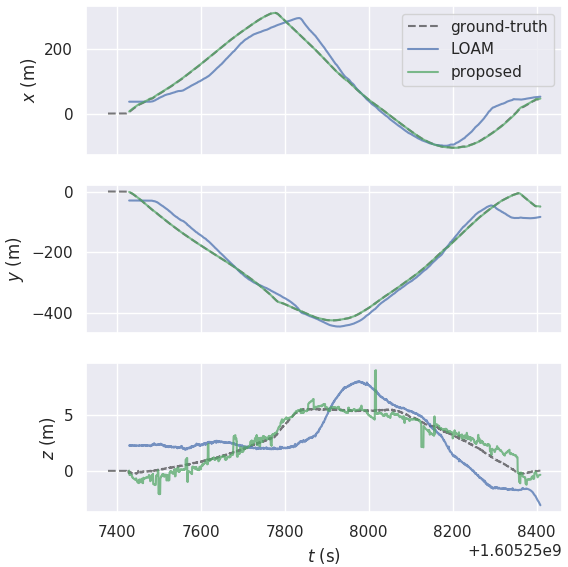}}
  \caption{(a) Trajectories of the ground truth, LOAM and the proposed long-term localization system. (b) Position errors along x-y-z axes within a time window.}
  \label{fig:Result of long-term localization}
\end{figure}

Since LOAM only relies on the LiDAR odometry, the RMSE was relatively large because of the accumulated error of the odometry. 
For the proposed method, the relocalization algorithm based on the semantic cluster map was used to correcting the accumulated error, and the position estimation between frames of the laser odometry is accurate, and therefore the RMSE of the proposed long-term localization system is smaller. 
The trajectories of ground truth, LOAM and the proposed method are shown in Fig.\ref{fig:Result of long-term localization}(a) by using the dataset 2020-11-05 as an example, and the x,y,z position error are shown in Fig.\ref{fig:Result of long-term localization}(b).

\section{Conclusion}
A novel relocalization method based on semantic cluster map is proposed to realize high accuracy relocalization and real-time localization in long-term environments. 
A robust semantic cluster map is built by extracting pole-like objects from raw 3D LiDAR points to resolve the long-term challenge.
A semantic cluster association algorithm on the basis of geometric consistency vertification is proposed to obtain the pose transformation between the local cluster map and the global cluster map.
And a long-term and real-time localization system is developed based on the relocalization module and LiDAR odometry.
Finally, several experiments are performed to illustrate the effectiveness of the proposed approach.
In the future work, we will extend to utilize more objects in long-term environments and make it more general rather than relying heavily on pole-like objects.



\bibliographystyle{unsrt}
\bibliography{ref.bib}
\end{document}